\documentclass[referee]{sn-jnl}

\usepackage[superscript,biblabel]{cite}
\usepackage{graphicx}%
\usepackage{multirow}%
\usepackage{amsmath,amssymb,amsfonts}%
\usepackage{amsthm}%
\usepackage{mathrsfs}%
\usepackage[title]{appendix}%
\usepackage{xcolor}%
\usepackage{textcomp}%
\usepackage{manyfoot}%
\usepackage{booktabs}%
\usepackage{algorithm}%
\usepackage{algorithmicx}%
\usepackage{algpseudocode}%
\usepackage{listings}%

\usepackage{bm}

\usepackage{enumitem}  
\usepackage{relsize}
\usepackage{amsthm}
\usepackage{bbm}
\usepackage{lineno}
\usepackage{bigints}

\DeclareMathOperator*{\argmin}{arg\,min}

\begin{document} 

\title{Rhythmic sharing: A bio-inspired paradigm for zero-shot adaptive learning in neural networks}


\author[1]{\fnm{Hoony} \sur{Kang}}\email{kang.hoony.2@gmail.com}

\author[1,2]{\fnm{Wolfgang} \sur{Losert}}\email{wlosert@umd.edu}

\affil[1]{\orgdiv{Department of Physics}, \orgname{University of Maryland},  \orgaddress{\city{College Park}, \postcode{20740}, \state{MD}, \country{USA}}}

\affil[2]{\orgdiv{Institute for Physical Science and Technology}, \orgname{University of Maryland},  \orgaddress{\city{College Park}, \postcode{20740}, \state{MD}, \country{USA}}}


\date{}


\abstract{
  
The brain rapidly adapts to new contexts and learns from limited data, a coveted characteristic that artificial intelligence (AI) algorithms struggle to mimic. Inspired by the mechanical oscillatory rhythms of neural cells, we developed a learning paradigm utilizing link strength oscillations, where learning is associated with the coordination of these oscillations. Link oscillations can rapidly change coordination, allowing the network to sense and adapt to subtle contextual changes without supervision. The network becomes a generalist AI architecture, capable of predicting dynamics of multiple contexts including unseen ones. These results make our paradigm a powerful starting point for novel models of cognition. Because our paradigm is agnostic to specifics of the neural network, our study opens doors for introducing rapid adaptive learning into leading AI models.
}

\maketitle

\section{Introduction} 

Much of the dynamics that govern the natural world experience drifts in internal parameters, leading to non-stationary dynamics with statistics that change in time. Examples are changes in atmospheric chemistry, tectonic plate movement, or the transition of a tumor's state from benign to malignant. The effects of nonstationarity can be dramatic enough to alter its entire probability space, rendering simple tools such as mean-centering of the data insufficient. Although the initial impact of system parameter changes is often barely discernible in the data, detecting them may allow us to anticipate more severe long-term consequences such as climate change, earthquakes, or metastases. Therefore, the classification and prediction of dynamics that experience parameter drifts have garnered considerable attention across disciplines.

While machine learning has been proven to be a powerful tool in predicting the dynamics of complex systems in the absence of mathematical models, most methods optimize prediction for tasks originating from the \textit{same} context, i.e., stationary probability distribution. Learning is implemented by having the link strengths of the network monotonically increase or decrease during training, until an optimal configuration is found\cite{backprop,hebbian}. Furthermore, to predict nonstationary data, one must first classify the different states present in the data, and then feed these contextual tokens into the machine so that it can understand how the context of the training data changes over time.  Here, we define `states' as sets of trajectories that belong to the same stationary probability distribution, such as those of an attractor or some other set with invariant long-term statistical properties. Because each state represents a distinct stationary probability distribution, a state can be considered the underlying `context' of the time series. In the absence of these tokens, the machine cannot make sense of the different probability spaces in the data and tends to forget data whose context clashes with another's (catastrophic forgetting \cite{catastrophic1,catastrophic2}). The brains of living organisms excel at rapidly sensing subtle contextual changes in their environment, and do not suffer from catastrophic forgetting. 

\subsection{A new learning paradigm}
Inspired by the mechanochemical interactions of astrocytes and the neuronal synapse, we conjecture a new paradigm of neuroplasticity.  Recent work has suggested that learning may be implemented with biomechanics, since squeezing of neuronal synapses strengthens synaptic transmission \cite{ucar}. Our learning paradigm is inspired by this potential role of biomechanics and harnesses it in two parts.

\textit{First,  we propose that learning involves rhythmic variations in link strength.}
This is inspired by the recent findings of our team and others that biomechanics exhibits spontaneous rhythms \cite{qixin_elife,qixin_pnas, abby, actincalciumosc}.

\textit{Second, we propose that learning occurs via coordination of the phases of these rhythmic variations.} Astrocytes inspire this key element of our algorithm: Astrocytes, a type of glia abundantly present in the brain that wrap around thousands of synapses, exhibit rhythmic hotspots of biomechanical activity, which may enable the coordination among synapses \cite{kate}.  

\begin{figure*}
    \includegraphics[width=\textwidth]{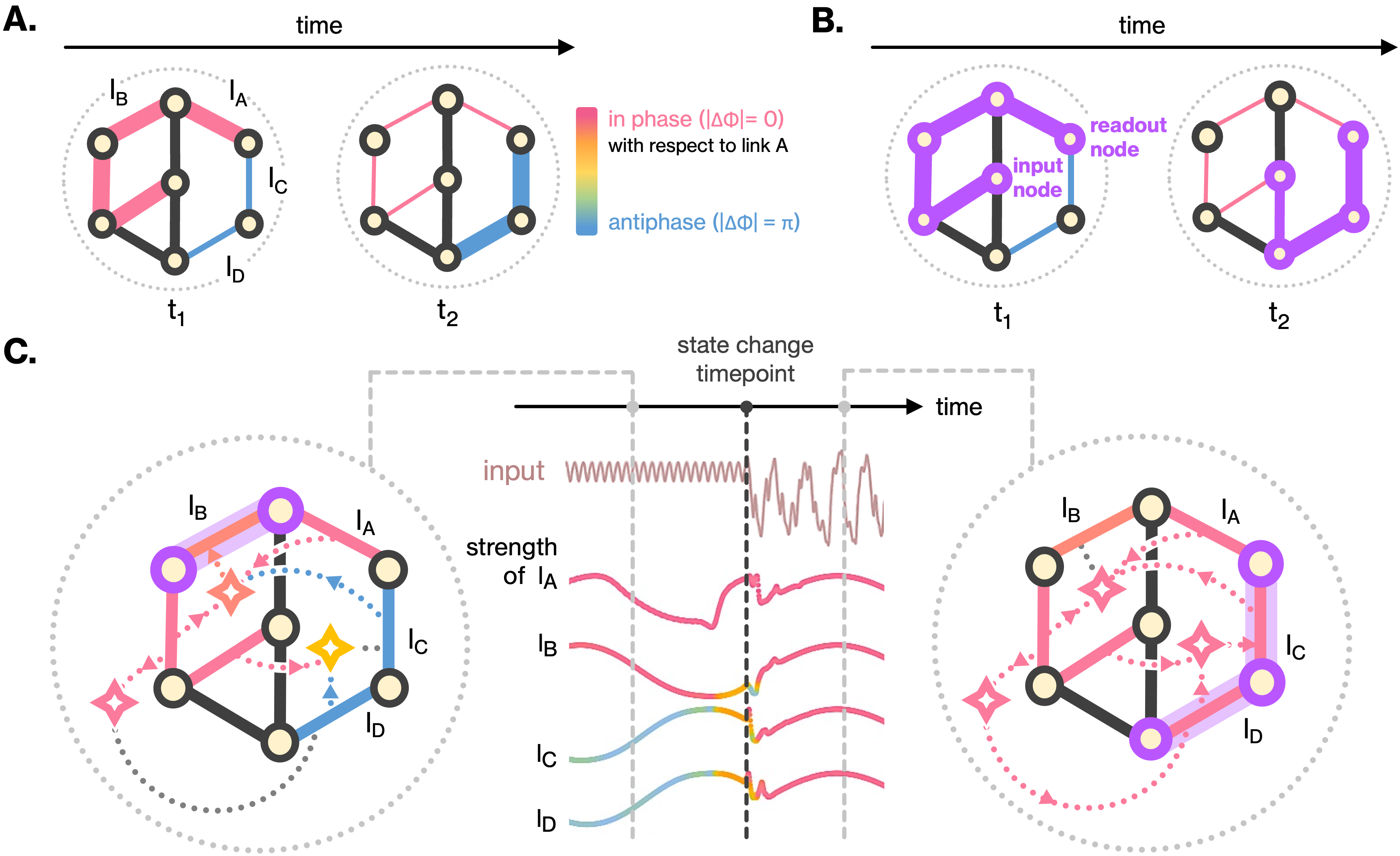}
        \caption{\textbf{Schematic of rhythmic sharing.} A. Temporal snapshots of oscillating link strengths, indicated by their widths. In this example, the degree of oscillation coordination, i.e., synchrony, is fixed, meaning the relative phases between links are fixed. B. The effect of oscillating link strengths on information flow through the network. At each timepoint, the current value of the mean phase alters which links are the strongest, rerouting how information flows through the network. As an example, the most coherent path of information between some input node to some readout node is shown in purple. C. Changing input alters the subset of active links in the network, altering the coordination among link oscillations. At earlier times, $l_B$ was activated by the periodic orbit input, encouraging it to synchronize with the mean phase of its connected links (the star), which is mostly still pink. Upon a change of input to a chaotic state, $l_C$ and $l_D$ become active, and they synchronize to their mean fields. This results in links $l_C$ and $l_D$ becoming synchronized to $l_A$. However, $l_B$, no longer active, is unaffected by its mean field.}
    \label{fig:fig1}
\end{figure*}

These two concepts work together to generate a powerful learning paradigm: We implement these rhythms as slow oscillations in link strength of some or all links, and allow their phases to be different for each link (fig. \ref{fig:fig1}A). Therefore information flowing through the network can utilize multiple subnetworks, depending on the phases of the oscillations at a given time (fig. \ref{fig:fig1}B).

The second part of the paradigm is that the phases of each link can change depending on the information it processes, e.g., different states, as illustrated in fig. \ref{fig:fig1}C. We propose that links that are active in processing a given state can adjust their phases rapidly to synchronize to other links. This synchronization is mediated by objects that can integrate information from subsets of links (inspired by the integrative role of astrocytes and denoted as stars in fig. \ref{fig:fig1}C).  

Because input of a different state results in active link changes, state changes manifest as collective changes to synchronization of the link phases as illustrated in fig. \ref{fig:fig1}C.  Therefore, link phase synchronization also functions as a classification token of each state.

We implement our algorithm on an artificial neural network and demonstrate that it can rapidly sense dynamic state changes in a range of dynamical systems including key model systems of chaos, and serve as a digital twin for prediction and state targeting.

\section{Results}
\subsection{Unsupervised state estimation with rhythmic sharing}

\begin{figure*}
    \includegraphics[width=\textwidth]{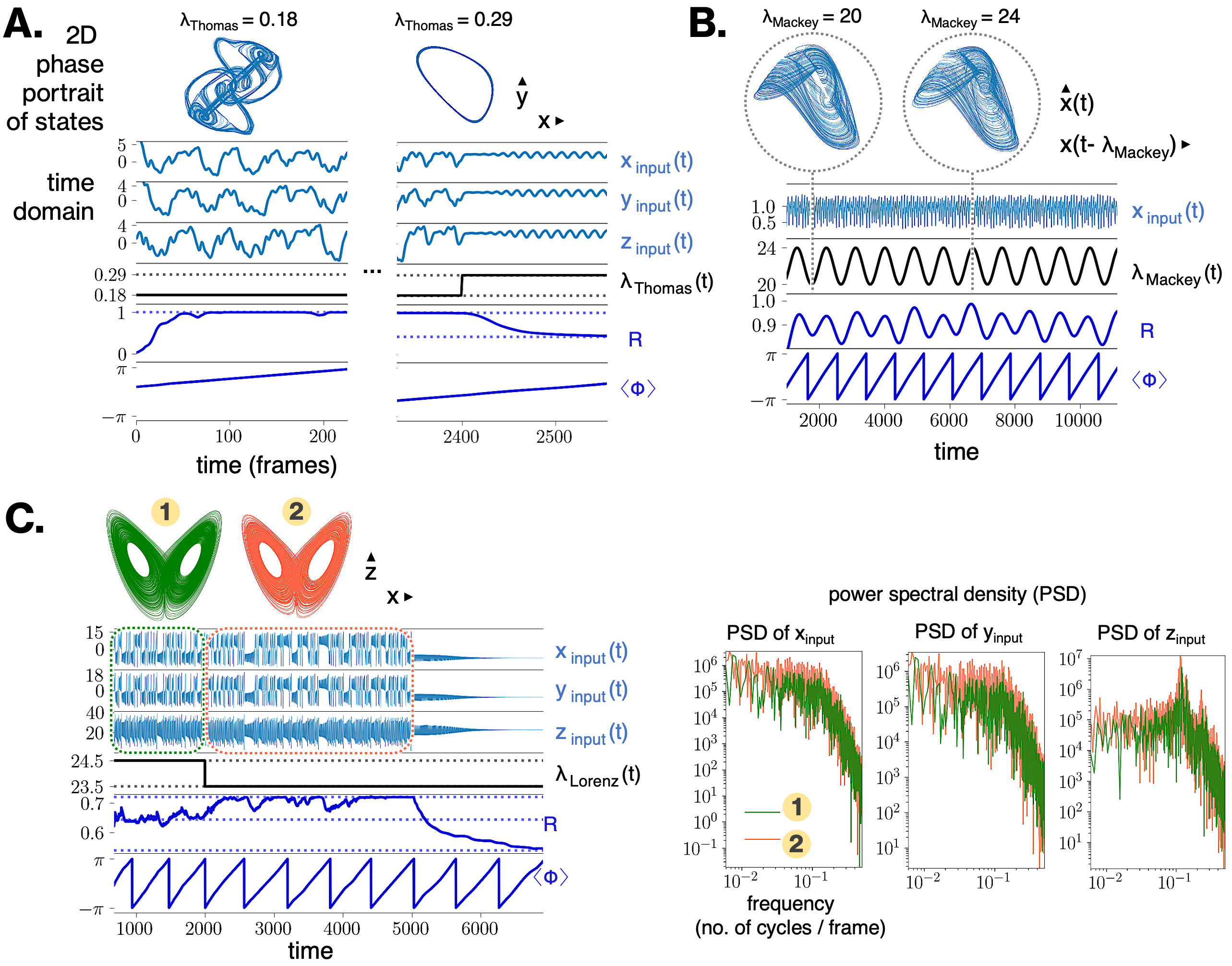}
        \caption{\textbf{Rapid sensing of dynamic state changes} A. A simulated trajectory of the 3D Thomas system (whose state variables are $x,y,z$) in orange, with the value of its internal parameter $\lambda_\text{Thomas}$ that was used to simulate the trajectory shown in purple. After an initial transient, $R$ remains near an equilibrium value of 1 for the entire chaotic trajectory. When $\lambda_\text{Thomas}$ is abruptly switched to $0.29$, the trajectory switches from a chaotic state to a periodic orbit. $R$ senses this switch immediately and settles to a new equilibrium value. B. The same procedure conducted for the Mackey-Glass system, whose internal parameter $\lambda_\text{Mackey}$ was changed sinusoidally during simulation. C. (Left) $R$ for the 3D Lorenz system. $R$, although noisier than the Thomas system, evolves to a different equilibrium value at the exact time $\lambda_\text{Lorenz}$ was switched, falling to a different value as the system transitions to a dead signal. (Right) The power spectral densities (PSD) of the time series of each state variable, corresponding to when the system is at stable chaos (green) and unstable, or transient, chaos (orange). That their spectral differences are also not obvious contributes to the difficulty in sensing this state transition.}
        \label{fig:fig2}
\end{figure*}

\subsubsection{Detecting simple state changes}

We choose the 3D Thomas' cyclically symmetric system\cite{thomas} as our first example input because it hosts numerous states with different values of its internal parameter, $\lambda_{\text{Thomas}}$. 

We simulate data where $\lambda_{\text{Thomas}}$ starts at $0.18$, which results in a chaotic trajectory, and then suddenly switches to $0.29$, which results in a periodic orbit. This data is then sent frame-by-frame into the nodes of a neural network whose connection weights and link phases are initially random. The evolution of the link phases as data enters the neural network is then tracked, as shown in fig. \ref{fig:fig2}A. The collective dynamics of the link phases are characterized by their order parameter $R \in [0,1]$ and the mean phase $\langle \Phi \rangle\in [-\pi,\pi)$. $R(t)$ measures the degree of global synchronization of link phases: $R=0$ indicates uniform dispersion, i.e., complete decoherence, of the phases, while $R=1$ indicates complete synchronization of the phases.

We emphasize two observations about $R(t)$. First, $R$ converges to some equilibrium value whenever the input trajectory belongs to a stationary state. For instance, $R$ evolves from its initial condition $0$ and reaches a steady state value $1$ within roughly two `cycles' of the chaotic state ($\lambda_{\text{Thomas}}=0.18$) and maintains this value hereafter. This means that the links detected an invariant measure underlying the trajectory, even when the trajectory is chaotic; in other words, the links detected that a stationary probability distribution underlies the time evolution of the trajectory. Second, each state gives rise to a unique equilibrium value of $R$. Therefore, the evolution of $R$ mimics the evolution of $\lambda_{\text{Thomas}}$.

The dynamics of the mean phase $\langle \Phi \rangle(t)$ are not as interpretable yet, as they do not respond uniquely to how $\lambda_{{\text{Thomas}}}$ evolves in time. However, they will be critical in the ability of the network to recall and predict various states (section \ref{recall}).

\subsubsection{Sensing continuous parameter changes on systems with memory}

The future evolution of a system may have a nontrivial dependence on its history. These systems are particularly abundant in biology, such as the immune and nervous systems\cite{ddeimmune,ddenervous}. One such example is the 1D Mackey-Glass system\cite{mackey}. For this system, the internal parameter $\lambda_{\text{Mackey}}$ that induces state changes represents the time delay, or memory, of the dynamics. We vary this parameter in a sinusoidal manner when simulating data.

Fig. \ref{fig:fig2}B shows the response of the links to this input data. While $R$ is shakier, likely due to the system's complexity and the continuous nature of the parameter change, it still successfully mimics the evolution of $\lambda_{\text{Mackey}}$.

\subsubsection{Detecting warning signs of catastrophic phenomena}\label{lorenz}

Finally, we study a system where changes to its internal parameters are not immediately apparent or measurable with current methods, but whose detection is critical to prevent the impending collapse of a system's dynamics.

A well-known chaotic system that exhibits this behavior is the 3D Lorenz system\cite{lorenz}. Its internal parameter $\lambda_{\text{Lorenz}}$ is responsible for state changes. The system exhibits a chaotic trajectory when $\lambda_{\text{Lorenz}}=24.5$. When $\lambda_{\text{Lorenz}}$ abruptly changes to $23.5$, the chaotic trajectory shifts into another similar, but transient chaotic trajectory until it suddenly collapses into a dead signal after some probabilistic time\cite{yorke}. Indeed, fig. \ref{fig:fig2}C shows the close similarity in the spectral properties of the two chaotic trajectories; to resolve the spectral peak differences ($\approx$ 0.1-0.2 Hz), at least 50-100 cycles of the frequency bands containing the peaks must be observed.

The behavior of $R(t)$ of fig. \ref{fig:fig2}C demonstrates that the links sensed the moments when the states were switched: from the first chaotic trajectory to the chaotic transient, and then to the dead signal. The time it took for $R(t)$ to escape its range of values while it was sensing the chaotic trajectory and to enter its new range associated with the chaotic transient is around 100 frames, or 11 cycles.

\begin{figure*}
    \includegraphics[width=\textwidth]{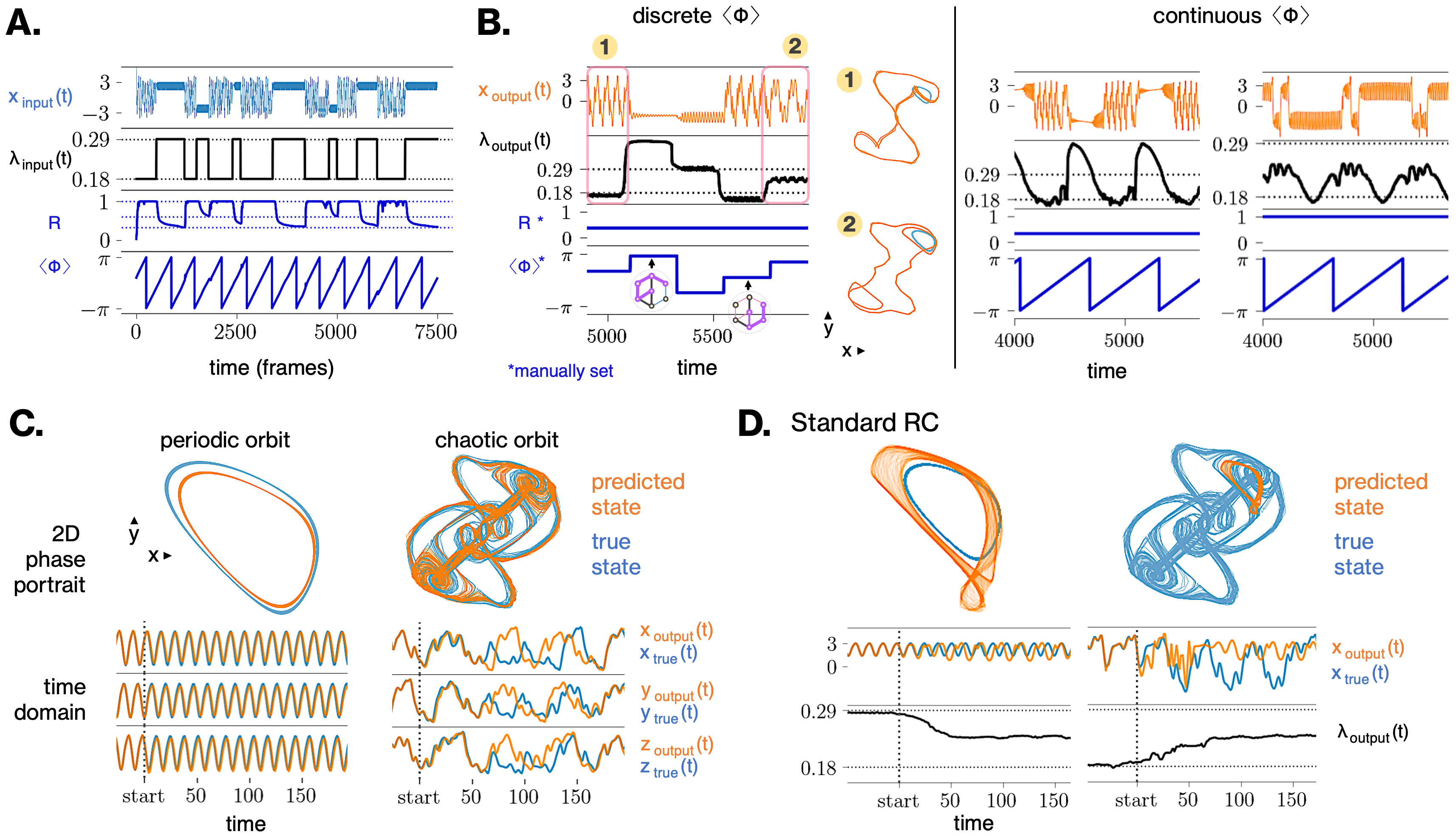}
        \caption{\textbf{Generating a digital twin for prediction and state targeting} A. Training data used for the Thomas system and the response of the links, characterized by $R$ and $\langle \Phi \rangle$. B. Closed-loop predictions of different states by modifying $R$ and $\langle \Phi \rangle$ in real time. Some extrapolated states that the network has never seen before are highlighted with pink boxes in the zoomed-in time series, and their states are respectively shown in coordinate space in orange as well. Here, the light blue state represents the initial input state (the periodic orbit), before closed-loop prediction began. Each value of $R$ and $\langle \Phi \rangle$ amounts to sampling different network paths. C. Closed-loop prediction of individual states present in the training data without being contaminated by other states. The phase portrait representation (top) shows the state of the predicted trajectory that ran for over 5000 frames, which matches closely to the state of the true trajectory. A zoomed-in, time-domain representation of the prediction (bottom) showcases the forecasting accuracy. `Start' represents the beginning of the closed-loop prediction stage. D. Closed-loop time series predictions of a standard RC, which learned the training data as one state and therefore could not predict each state independently.}
        \label{fig:fig3}
\end{figure*}

\subsection{State transformation and recall by rerouting information flow with rhythmic sharing}\label{recall}

A powerful feature of a neural network is its ability to store multiple dynamical states. If states were tagged with contextual tokens in the training data, one may recall a specific state by inputting the associated token entry. Without these tokens, recall is very difficult.

While $R$ seems to fulfill the role of the required contextual token, the link phases' mean dynamics are also governed by $\langle \Phi \rangle$, both of which shape the `neural representation' of the input. Therefore, we cannot neglect its specification when attempting to recall a state. Yet the latter's role is not visually clear from fig. \ref{fig:fig2}, as it does not seem to respond to state changes. To assess this, we will observe how the neural output changes in response to changes to $R$ and $\langle\Phi\rangle$. 

To obtain an output from the network, we train the network as a reservoir computer (RC), a form of a recurrent neural network that has been noted to perform well for chaotic dynamics prediction \cite{jaeger_rc,maass_lsm,chaos,chaos_neuron}. The training of an RC involves finding the optimal combination of nodes such that their combined output best recovers the true value of the input time series at the next timestep.

We reuse the Thomas system, as its large repository of states is well suited for our demonstration of state targeting. We train the RC with simulated data that alternates between two values of $\lambda_\text{Thomas}$, with the goal being to predict the dynamics of each state independently (fig. \ref{fig:fig3}A). We will henceforth refer to the $\lambda_\text{Thomas}$ of the input, or training, data as $\lambda_{\text{input}}$ to clarify that this is the internal parameter of the input trajectory $(x_\text{input}(t),y_\text{input}(t),z_\text{input}(t))$ that is sent to the network. Our goal is motivated by the fact that if the RC can separate and learn stationary events differently with the assistance of rhythmic sharing, it must also be capable of predicting each stationary event indefinitely without mixing the other events that it has seen during training. The alternations between the states are made aperiodic to ensure our results are not due to some resonance between the frequency of the link oscillations and that of $\lambda_{\text{input}}(t)$. From fig. \ref{fig:fig3}A, we see that, as before, $R(t)$ tracks the state changes in the data: $\lambda_{\text{Thomas}}=0.18$ admits one chaotic state, and $\lambda_{\text{Thomas}}=0.29$ admits two coexisting periodic orbits, resulting in three equilibrium values of $R$.

Once the optimal set of output nodes is obtained, we run the network in a closed-loop prediction scheme, where the network output (the prediction of the next frame) serves as the next input for the prediction of the subsequent frame. We freeze $R$ once it reaches its equilibrium value for the input state, then commence the closed-loop prediction scheme. Then, as the closed-loop prediction is running, we manually change  $\langle \Phi \rangle$ to observe its effects on the output. 


Fig. \ref{fig:fig3}B showcases three different examples of how $\langle \Phi\rangle(t)$ was changed during the closed-loop prediction. In the first example where $\lambda_\text{input}=\lambda_\text{{periodic orbit}}$ as $\langle \Phi \rangle(t)$ evolves discretely, each discrete value of $\langle \Phi \rangle$ results in neuronal output of a different state of the Thomas system in the neighborhood of $\lambda_\text{{periodic orbit}}$. Most of these states map to values of $\lambda_{\text{Thomas}}$ that the network never encountered during training. Because each $\langle \Phi\rangle$ value qualitatively represents a different static weight configuration of the network, one may say that we are effectively reading out different states embedded in different information pathways associated with each configuration.

In the second example, where $\lambda_{\text{input}}=\lambda_\text{{periodic orbit}}$ but the mean phase $\langle \Phi \rangle$  evolves \textit{continuously} in a linear manner, it is evident that $\lambda_{\text{output}}(t)$ (the internal parameter associated with the output trajectory $(x_\text{output}(t),y_\text{output}(t),z_\text{output}(t))$) is entrained with $\langle \Phi \rangle$. This demonstrates that $\langle \Phi \rangle$ is an effective controller that can reliably steer the neural states to desired trajectories corresponding to various $\lambda_{\text{output}}$; in other words, accessing different paths of the network recalls different states. These observations hold for the third example, where setting $\lambda_{\text{input}}=\lambda_\text{{chaos}}$ yields a different extrapolation curve in the neighborhood of $\lambda_{\text{chaos}}$. The results suggest that the mean-field phase parameters can be used to predict how the time series will evolve when the context, or state, of the time series switches in real-time---in other words, a digital twin.

With these findings, we may now easily achieve our original goal:  by freezing the proper values of $R,\langle \Phi \rangle$ such that $\lambda_{\text{output}}=\lambda_{\text{input}}$, the network can predict the time series of each state present in the training data separately without contamination, as shown in fig. \ref{fig:fig3}C. Although its predictive performance is not as accurate as an RC trained purely on either state alone, a single RC endowed with rhythmic sharing can accurately predict the long-term dynamics of each individual context or state.

In contrast, a standard RC not equipped with contextual tokens cannot learn the different states in the nonstationary training data as distinct events, and attempts to treat the data as originating from one \textit{stationary} event. This can be seen in fig. \ref{fig:fig3}D, where the RC predicted a trajectory whose $\lambda_{\text{output}}$ is intermediate of the two $\lambda_{\text{input}}$ of the training data. That the reservoir treated the events in the training data as one is evident by the fact that it outputs the same trajectory for both test states.

\section{Conclusion}

Motivated by neuro-glial interactions of the brain, we developed a learning paradigm consisting of two key elements: the slow oscillations of the links and the ability to change their coordination. We implemented our model into an artificial neural network to probe the role of these mechanisms in neural representation.

In our model, links rapidly adjust their synchronization to identify the current state dynamics. Thus, the synchrony of links identifies the current state of the dynamics in the data. By self-adjusting link strength rhythms for different states, the network is capable of learning nonstationary dynamics. This is a powerful feature that regular machine learning algorithms cannot handle without user-provided contextual tokens, despite the ubiquity of nonstationary dynamics in real-world data. Our results demonstrate that the network is capable of rapidly identifying anomalous events in data and signaling warning cues for impending disasters. Furthermore, this is achieved in real time because the network did not require prior training. 

 Slow link strength rhythms enable extrapolation of dynamics of numerous unseen states, and by training the network endowed with rhythmic sharing, we can predict the stationary dynamics of each state. This asset may be used to harness the network as a digital twin, capable of adapting its prediction when the physical twin experiences state changes.

We restricted the application of our algorithm to systems whose nonstationarity is induced by a single parameter. This is because state changes brought on by more than one parameter cannot be causally differentiated from the $R(t)$ plot alone, which will combine all sources of change to the system. A future direction would be to design a model where multiple clusters can individually track the nonstationary evolution of different internal parameters of a system, thereby creating small-world networks of links.

\section{End Matter}

\subsection{The input systems}
\subsubsection{Thomas' cyclically symmetric system}
The following equations govern the 3D Thomas system\cite{thomas}:
\begin{align}
\begin{split}
    \dot{x}=\sin(y)-bx\\
    \dot{y}=\sin(z)-by\\
    \dot{z}=\sin(x)-bz
    \end{split}
    \label{eq:thomas}
\end{align}

The parameter $b$ induces bifurcations in the system, and is recast as the parameter $\lambda_{\text{Thomas}}$ in the main text.

\subsubsection{Mackey-Glass system}

The normalized form of the 1D Mackey-Glass system is given by\cite{mackey}:
\begin{equation*}
\dot{x}(t)=\beta \frac{x(t-\tau)}{1+x(t-\tau)^n}-\gamma x(t)
    \label{eq:mackey}
\end{equation*}

Here, the exact time dependence of the state variable $x$ is provided for clarity. $\tau$ is the time delay parameter responsible for inducing bifurcations of the attractor (parameterized with phase space coordinates   $(x(t),x(t-\tau)$), and is recast as the parameter $\lambda_{\text{Mackey}}$ in the main text. We fix $\beta = 0.2, \gamma = 0.1$, and $n=10$, while we oscillate $\tau$ sinusoidally between $\tau=20$ and $\tau = 24$ with a slow angular frequency ($= 0.007$) during the simulation of our data.

\subsubsection{Lorenz system}
The following equations govern the 3D Lorenz system\cite{lorenz}:
\begin{align}
\begin{split}
\dot{x}&=\sigma (y-x)\\
\dot{y}&=x(\rho-z)-y\\
\dot{z}&=xy-\beta z
\end{split}
\label{eq:lorenz}
\end{align}

Here, the parameter $\rho$ is responsible for bifurcations of the system and is recast as the parameter $\lambda_\text{Lorenz}$ in the main text. We fix the other parameters to values $\sigma = 10$, $\beta = 8/3$ for the duration of the entire simulation, while we switch the value of $\rho$ abruptly as in the case for the Thomas system. For $\rho\gtrapprox 24.06$, the system permits a globally stable strange attractor that coexists with two stable fixed points, but the former's trajectories never hit the latter's basin boundary. For $13.93 \lessapprox \rho\lessapprox 24.06$, the strange attractor turns into a chaotic saddle, which also coexists with the fixed points, but trajectories of the saddle will hit the saddle boundary after some probabilistic time\cite{yorke}.

We switch $\rho$ from 24.5 to 23.5 in our simulated data. Our choice of $\rho=23.5$ is due to its proximity to the bifurcation threshold at $\rho \approx 24.06$. By minimizing this difference, we minimize visual changes to the trajectory as the chaotic attractor transitions into a saddle.

\subsection{Rhythmic sharing}
We model the link interactions with the following governing equation:

\begin{align}
    \frac{d\boldsymbol\Phi}{dt}=\boldsymbol\omega_0+(\varepsilon_1+\varepsilon_2\mathbf{\hat{Q}}^\text{T}\mathbf{n}^*)\circ\sin(\boldsymbol\Psi-\boldsymbol\Phi+\boldsymbol{\gamma})
    \label{eq:phase}
\end{align}

\noindent where $\circ$ denotes the Hadamard product. $\boldsymbol\Phi$ is a vector of phases $[\phi_1\,...\,\phi_{N_l}]^\text{T}$ with $N_l$ being the total number of links. $\boldsymbol\omega_0$ is a vector of initially imposed natural frequencies of the links. The vector of \textit{local} mean fields $\boldsymbol\Psi(t)\in \mathbb{R}^{N_l}$ that each link is coupled to is defined by
\begin{align}
    \mathbf{r}(t)\circ e^{i\boldsymbol\Psi(t)}=\mathbf{\hat{A}}_{\boldsymbol\Phi}e^{i\boldsymbol\Phi(t)}
\end{align}

\noindent where $\mathbf{r}\in \mathbb{R}^{N_l}$ is a vector of \textit{local} order parameters whose elements describe the degree of synchronization across the particular local mean field that each link is coupled to. $e^{i\cdot}$ is taken to operate on $\cdot$ element-wise, e.g., $e^{i\boldsymbol\Psi(t)}=(e^{i\psi_1},e^{i\psi_2},...,{e^{i\psi_{N_l}})}$. $\mathbf{\hat{A}}_{{\boldsymbol\Phi}_{ij}}\equiv \mathbf{A}_{{\boldsymbol\Phi}_{ij}}/|\mathbf{A}_{{\boldsymbol\Phi}_{i\bullet}}|$, where $\mathbf{A}_{\boldsymbol\Phi}\in \mathbb{R}^{N_l \times N_l}$ is a random phase adjacency matrix, binary for simplicity, whose elements equal 1 if the links are connected and 0 otherwise, and the normalization factor $|\mathbf{A}_{{\boldsymbol\Phi}_{i\bullet}}|$ is the $L^1$ norm of the $i$-th row of $\mathbf{A}_{\boldsymbol\Phi}$. The density of non-zero elements of $\mathbf{A}_{\boldsymbol\Phi}$ matrix is a tuneable hyperparameter. We note in passing that while we limit the focus of this study to 1:1 synchronization, arbitrary orders of $p:q$ synchronization (for clusters) may be readily implemented by modifying the argument of the right-hand side of the eq. \ref{eq:phase} by taking $\sin(\boldsymbol\Psi-\boldsymbol\Phi)\rightarrow\sin(\mathbf{p}\circ\boldsymbol\Psi-\mathbf{q}\circ\boldsymbol\Phi)$, where $\mathbf{p}$ and $\mathbf{q}$ are vectors of positive integers whose elements $\mathbf{q}_i,\,\mathbf{p}_i$ denote the desired synchronization order between the phase $\boldsymbol{\Phi}_i$ and its connected mean field  $\boldsymbol{\Psi}_i$. However, careful engineering of the structure of $\mathbf{A}_{\boldsymbol\Phi}$ will most likely be needed to minimize competition of different synchronization orders between the links. Finally, $\boldsymbol{\gamma}$ is a phase-lag vector (treated as another hyperparameter) whose role is akin to that of the Sakaguchi-Kuramoto model in promoting asymmetric communication\cite{sakaguchi}.


The amplitude of the coupling, $\varepsilon_1+\varepsilon_2\mathbf{\hat{Q}}^\text{T}\mathbf{n}^*$, encodes the causal direction of nodal activity promoting link interactions. Here, the elements of the normalized incidence matrix $(\mathbf{\hat{Q}}^\text{T})_{ij}\equiv(\mathbf{Q}^\text{T})_{ij}/|(\mathbf{Q}^\text{T})_{i\bullet}|$ are of the incidence matrix $\mathbf{Q}\in\mathbb{R}^{N_n\times N_l}$, whose numerator elements equal 1 if a link is incident to a node (self-loops included) and 0 otherwise, and the normalization factor $|(\mathbf{Q}^\text{T})_{i\bullet}|$ is the $L^1$ norm of $\mathbf{Q}^\text{T}$ across the $i$-th row. We rescale the node states $\mathbf n \in (-1,1)$ to $\mathbf{n^*}\in (0,1)$ by defining $\mathbf{n}^* \equiv (\mathbf{n}+\mathbbm{1}_{N_n})/2$ (where $\mathbbm{1}_{N_n}=(1,1,..,1) \in \mathbb{R}^{N_n}$ denotes a vector of 1s of size $N_n$), because lack of activity of the $i$-th node, $\inf(\mathbf{n}_i)= -1$, should translate to an \textit{absence} of link-to-link interaction, $\inf(\mathbf{n}_i^*)= 0$. The hyperparameter $\varepsilon_1$ can engineer the disparity in phase coherence between different states. However, it is not to be set to a nonzero value if the hyperparameter $\varepsilon_2=0$ as this will promote global synchronization or desynchronization across the link network asymptotically regardless of the state of the system.

Finally, we discuss the vector hyperparameter of slow natural frequencies $\boldsymbol\omega_0$. While its distribution and values are not constrained a priori, we consider the simple binary case where a set percentage of links share the same natural frequency $\omega_0\neq 0$, itself a hyperparameter, and the rest are set to 0. We denote the density of links with nonzero natural frequencies as $\Lambda(\omega_0)\in (0,1]$, which is also a hyperparameter.

The \textit{scalar} global order parameter used to track the coordination of link phases is defined similarly, $R(t)\in[0,1]$,  which is the degree of global synchrony across \textit{all} the link phases:
\begin{align}
  R(t) e^{i\langle \Phi \rangle (t)}=\frac{1}{N_l}\sum_{k=1}^{N_l}e^{i\boldsymbol\Phi_k(t)}  
\end{align}

Here, $\langle \Phi \rangle(t)$ is the corresponding \textit{scalar} global mean phase. We note that this is equivalent to the definition of the local mean fields defined previously, except where all elements of $\mathbf{A}_{\boldsymbol\Phi}$ are replaced with 1.

The equation describing the evolution of the node states of our recurrent neural network (a type of reservoir computer called an echo state network\cite{jaeger_rc,jaeger_esn}) $\mathbf{n}\in \mathbb{R}^{N_n}$ is
\begin{align}
\begin{split}
       \mathbf{n}(t+\Delta t)=\,\alpha\mathbf{n}(t)+(1-\alpha)\tanh\big[\mathbf{A}\mathbf{n}(t)+\mathbf{W}_{\text{in}}\mathbf{u}(t)+\boldsymbol\xi\big] 
\end{split}
    \label{eq:n}
\end{align}

\noindent where the activation function $\tanh[\cdot]$ is applied to each vector equation component in its argument separately. It is to be noted that when we refer to a node, neuronal, or reservoir state, we mean a point of the vector $\mathbf{n}\in(-1,1)^{N_n}$; all other uses of the term `state' specifically refers to an invariant set, such as an attractor. $\alpha$ is the leakage hyperparameter that sets the rate of update and memory of $\mathbf{n}$, $\mathbf{A_n}\in\mathbb{R}^{N_n\times N_n}$ is another random adjacency matrix between the nodes (scaled by its spectral radius, a hyperparameter), $\mathbf{W}_{\text{in}}\in \mathbb{R}^{\text{N}_n\times \text{N}_u}$ is the random input matrix whose elements are chosen uniformly from the closed interval $[-\delta,\delta]$ where $\delta$ is the input scaling hyperparameter, $\mathbf{u}(t)\in \text{N}_u$ is the input, $\boldsymbol\xi\in \mathbb{R}^{N_n}$ is a node bias hyperparameter that can be used to shift the activation function of the neuronal nodes. While eq. \ref{eq:n} subsumes the nonlinearity of activation to each node, we note that it is equivalent to other recurrent neural network update equations up to a coordinate transformation\cite{coordinatechange}. As such, we do not lose generality by using an echo state-type network.

To use this network as a reservoir computer, the latent space of neuronal states is then projected back onto the original domain via a linear map $\tilde{\mathbf{u}}:\mathbb{R}^{\text{N}_n}\rightarrow \mathbb{R}^{\text{N}_u}$ where
\begin{align}
    \tilde{\mathbf{u}}(t)=\mathbf{W}_{\text{out}} \mathbf{n}(t)
\end{align}

As $\tilde{\mathbf{u}}(t+\Delta t)$ assumes the prediction of $\mathbf{u}_\text{data}$ after one timestep $\Delta t$, the goal is to obtain the matrix $\mathbf{W}_{\text{out}}$ such that $\tilde{\mathbf{u}}(t+\Delta t)$ best approximates $\mathbf{u}(t+\Delta t)$ for chosen hyperparameters. 

To incorporate oscillating link strengths with fixed relative amplitude to the neural network, eq. \ref{eq:n} may be modified by amending $\mathbf{A_n}$ to exhibit a simple oscillation dependence, $\mathbf{\tilde{A}_\mathbf{n}}(t)$, where
\begin{align}
    \mathbf{\tilde{A}_\mathbf{n}}(t)&\equiv\mathbf{A_n}\circ\bigg[1-\frac{m}{2}(1+\sin[\boldsymbol{\underline{\Phi}}(t)])\bigg]
\end{align}

Here, $m$ is the relative fraction of link strength change during oscillation. The purpose of constructing the modulation this way is so that the strength is always bound between the upper bound that is its original strength $\mathbf{A}_{\mathbf{n}_{ij}}$ and its lower bound $(1-m)\mathbf{A}_{\mathbf{n}_{ij}}$ to restrict the maximum spectral radius, such that the echo state property does not become inadvertently violated\cite{jaeger_esn}. Finally, the phase matrix $\boldsymbol{\underline{\Phi}}\in\mathbb{R}^{N_n\times N_n}$ is simply a collection of individual link phases ordered into the same structure as the adjacency matrix:
\begin{align}
    \begin{split}
      \boldsymbol{\underline{\Phi}}(t)_{ij}:=\left\{\begin{matrix}
    \underline{\phi}_{j\rightarrow i}&:\, \mathbf{A}_{{\mathbf{n}}_{ij}} \neq 0\\ 
    0 &:\, \mathbf{A}_{{\mathbf{n}}_{ij}} = 0
\end{matrix}\right.  
    \end{split}
\end{align}

As such, the vector $\boldsymbol\Phi$ in eq. \ref{eq:phase} is simply the non-zero elements of $\boldsymbol{\underline{\Phi}}$ collected into a vector of phases $[\phi_1\,...\,\phi_{N_l}]^\text{T}$ with $N_l$ being the total number of links, i.e., for some  $j,k\in [1,N_n]$, we can represent $\phi_i := \underline{\phi}_{j\rightarrow k}$ for all $i \in [1,N_l]$.

We note that we assume a 1-to-1 correspondence between the phase of link strength modulation and the phases of the links themselves. Therefore, the synchronization of the glial mediators is directly reflected in the neuronal dynamics as synchronized link strength modulations. While this was done mainly for simplicity, it was also experimentally observed that the oscillations of the type of chemical signals present in astrocytes may be 1:1 synchronized to the oscillations of the sort of biomechanical oscillations that are found in astrocyte hotspots\cite{actincalciumosc}. However, $f$ can take on other forms representing how the chemical dynamics of the astrocytes are translated to these mechanical waves.
 
While we formulated our model in the continuum limit $\Delta t \rightarrow 0$, we use first-order forward Euler discretization to simulate our model (and any modifications to it, as discussed in the following sections), with $\Delta t$ equal to that in eq. \ref{eq:n}.

\subsubsection{Oscillation death for isolated links}
	
In Kuramoto-like models\cite{kuramoto}, every oscillator is independently coupled to a subset of other oscillators in the network. If an oscillator is not coupled to any other oscillator, its phase will evolve with its unperturbed natural frequency $\omega_0$. 

However, our choice to couple each oscillating link to the mean field of its connections endows the system with the property that any isolated, active link will eventually quench its oscillations for $|\omega_0| < |\varepsilon|$. To see how oscillation death happens in our model, let us consider a single oscillator, $\phi$ (where we drop indices for brevity), that evolves without any link-to-link interaction:
\begin{align}
\dot{\phi}=\omega_0+\varepsilon \sin(\Psi_0-\phi)
\label{eq:death}
\end{align}

\noindent where $\Psi_0$ is the constant $\text{atan2}(0./0.)$, which is mathematically undefined yet computationally yields some numerical value due to floating point accuracy. $\varepsilon\equiv \varepsilon_1+\varepsilon_2 (\mathbf{Q^T\,n}^*)_\phi$ is the strength of the interaction for this particular oscillator, and which is bounded in general between $\varepsilon_1\pm\varepsilon_2$. Here, $\varepsilon$ is frozen as a constant for simplicity and to obtain an analytic solution.

By a translation, $\phi \rightarrow \phi -\Psi_0$, eq. \ref{eq:death} reduces to 
\begin{align}
 \dot{\phi}=\omega_0+\varepsilon \sin(-\phi)   
\end{align}

By use of a half-tan substitution, the solution can be easily shown to be:
\begin{align}
 \phi(t)=2\tan^{-1}\bigg[\frac{\varepsilon}{\omega_0}+\frac{\Delta}{\omega_0}\tan\bigg(\frac{\Delta}{2}t+\xi_0\bigg)\bigg]   
\end{align}
\noindent where $\Delta\equiv \sqrt{\omega_0^2-\varepsilon^2}$ and $\displaystyle\xi_0\equiv 2\tan^{-1}[-\frac{\varepsilon}{\Delta}+\frac{\omega_0}{\Delta}\tan\frac{\phi_0}{2}]$ with $\phi_0$ being the initial phase. It suffices to show that if the argument of the $\tan^{-1}$ has a real limit for large time given small $\omega_0$ (compared to the order of $\varepsilon$, which will be on the order of unity for synchronization), then $\tan^{-1}$ (and therefore $\phi$) will converge to a real value as $t\rightarrow \infty$.

For $|\omega_0| < |\varepsilon_0|$, $\Delta$ and $\xi_0$ are purely imaginary, making both the amplitude and the argument of the nested $\tan$ in the expression for $\phi(t)$ also purely imaginary. As such, $\phi(t)$ evolves as $2\tan^{-1}[a+b\tanh(c t + d)]$, where $a,\,b,\,c,\,d$ are real constants. Since $\tanh(c t + d)$ possesses a real limit for large $t$, $\phi(t)$ converges to a real limit, i.e.,the oscillation has quenched.

On the other hand, consider $|\omega_0|> |\varepsilon_0|$. Then $\Delta$ and $\xi_0$ are purely real, and $\phi(t)$ evolves as $2\tan^{-1}[a+b\tan(c t + d)]$, where $a,\,b,\,c,\,d$ are reused to denote arbitrary real constants. Because the argument does not have a real limit for large $t$, the phase $\phi(t)$ also does not. Therefore, $|\omega_0|< |\varepsilon|$ for oscillation death to occur in isolated links.

\subsection{Overview of reservoir computing with echo state networks}

This section provides a brief review of how training and predicting are conducted for an echo state network.

\subsubsection{Training}
First, a reservoir is initiated into a `warm-up phase' for $\tau_\text{warm}$ steps of $\Delta t$ of the input. As a reservoir computer has leaky memory, this stage is required to provide sufficient time for a reservoir to forget its initial state (usually a blank slate $\mathbf{n}=0$). Here, the input $\mathbf{u}$ is the training data $\mathbf{u}(t)=\mathbf{u}_\text{train}(t)$ so as to entrain $\mathbf{n}(t)$ to the dynamics of the input data without any feedback error. As this stage is used solely to calibrate the nodes' state space with that of the data, these reservoir states are discarded and not used for training.

Upon completion, the reservoir enters the `training phase' for $\tau_\text{train}$ steps of $\Delta t$. The now-calibrated states are once more entrained by $\mathbf{u}(t)=\mathbf{u}_\text{train}(t)$, but all the reservoir states during this phase $\mathbf{n}(t_0+\Delta t)$, $\mathbf{n}(t_0+2 \Delta t)$, ..., $\mathbf{n}(t_0+\tau_\text{train} \Delta t)$ are recorded. For notational brevity, $t_\text{warm}=\tau_\text{warm}\Delta t$ marks the end of the warm-up period. Using the concatenated state responses across time, the optimal output matrix $\mathbf{W}_{\text{out}}$ is then calculated via $L^2$ regression by minimizing:
\begin{align}
\begin{split}
   	\argmin_{\mathbf{W}_\text{out}}\bigg[\sum_{t=t_\text{warm}+\Delta t}^{t_\text{warm}+\tau_{\text{train}}\Delta t}||\mathbf{W}_{\text{out}}\mathbf{n}(t)-\mathbf{u}_\text{train}(t)||^2+\beta \,\text{tr}(\mathbf{W}_\text{out}\mathbf{W}_\text{out}^\text{T})\bigg] 
\end{split}
\end{align}

\noindent where  $\beta$ is the regularization hyperparameter.

\subsubsection{Predicting}\label{sss:prediction}

\noindent Equipped with the trained $\mathbf{W}_\text{out}$, we are now ready to commence the prediction phase. A  portion of the test data $\mathbf{u}_\text{test}$ is sent into the reservoir (which can either start from a blank slate again or from its last state during the training phase) for another warm-up period to forget its previous state and re-calibrate it to the new data to be predicted. We note in passing that the warm-up period for the test data need not be the same as that of training. The last frame of the test data in the warm-up period $\mathbf{u}_\text{test}(t_\text{warm})$ is now used as the initial condition to generate the first prediction, $\mathbf{\tilde{u}}(t_\text{warm}+\Delta t)$. This output is then fed back as the new input to obtain the reservoir's next state, forming a \textit{closed-loop} prediction for the next timestep. This process is iterated until the reservoir has advanced up to our desired prediction window.

\subsubsection{State targeting and prediction by steering $(R,\langle \Phi\rangle)$}

\noindent For our modified architecture, we perform the same procedure as above, but modify the warm-up stage of the prediction slightly if we wish to predict the time series of a single state, as it was done in fig. \ref{fig:fig3}C. As the network warms up to the test data, the links re-initialize their order parameter to the test state. Once the order parameter has reached its steady state (which we denote here as $R_0$), we freeze $R$ by imposing that the phases evolve in any arbitrary manner according to a prescribed vector-valued forcing $\mathbf{g}$ satisfying
\begin{align}
\begin{split}
    &\dot{\boldsymbol{\Phi}}(t)=\mathbf{g}(\boldsymbol{\Phi},t)\quad ,\quad t\geq t_{R_0}\\
    &\mathbf{g}(\boldsymbol{\Phi},t)=g(t)\mathbbm{1}_{N_l}\\
    &\boldsymbol{\Phi}(t_{R_0})=\mathbf{\Phi}_{R_0}
\end{split}
\label{eq:phasefreeze}
\end{align}

\noindent where $g$ is some scalar-valued function, $\mathbbm{1}_{N_l}=(1,1,..,1) \in \mathbb{R}^{N_l}$ denotes a vector of 1s of size $N_l$, $\mathbf{\Phi}_{R_0}$ denotes as the vector of phases at the first instance $t$ when $R=R_0$, which we denote as $t_{R_0}$. This is because the system of equations satisfies the constraint $R(t)=R_0$ for $t\geq t_{R_0}$, where $t_{R_0}<t_\text{warm}$, since the forcing does not depend on the phase, and so the Lyapunov exponent of the system is 0 (the phases do not diverge from each other in time).



 We choose $g=\Omega t$, with $\Omega = \omega_0$, such that the states can evolve slowly until the target mean phase $\langle \Phi \rangle=\langle \Phi\rangle_0$ can be obtained. Once our desired $\langle \Phi \rangle_0$ is reached, we replace $g=0$ to freeze the links permanently. We determine our desired $\langle \Phi \rangle_0$ by finding 
\begin{align}
\begin{split}
  	\langle \Phi \rangle_0 = \argmin_{\langle \Phi \rangle}\bigg[ &h(\mathbf{\hat{W}}_{\text{out}}\mathbf{n}(\boldsymbol{\Phi},t),\mathbf{u}_\text{test}(t))|\,\forall t \in [t_{R_0},t_{R_0}+\tau] \bigg] 
\end{split}
\label{eq:loss}
\end{align}

\noindent for $\tau > 2\pi/\Omega$ to ensure the space of static link strength configurations given $R_0$ is fully sampled during this process, and where $h$ is an error function of choice. The warm-up duration of the prediction is chosen such that all the targeting of the mean-field variables aforementioned happens before the end of the warm-up period, i.e.,$t_{R_0}+\tau \leq \tau_{\text{warm}}\Delta t$. For most systems, as we have used, a natural candidate is the root mean square error: $h^2=||\mathbf{\hat{W}}_{\text{out}}\mathbf{n}(\boldsymbol{\Phi},t)-\mathbf{u}_\text{test}(t)||^2$, but if the topology of the state does not change considerably with the system's bifurcation parameter or the system is contaminated with noise, an error function that can more finely resolve differences in states should be formulated and used. In summary, \\

\begin{enumerate}
    \item Evolve the phase with eq. \ref{eq:phase} with $\mathbf{u}_{\text{test}}$ until $R=R_0$.
    \item Replace the right side of eq. \ref{eq:phase} with eq. \ref{eq:phasefreeze} for a duration $\tau$ such that $R(t)=R_0$ while $\langle \Phi \rangle$ makes at least one full rotation during $\tau$. Identify the mean phase $\langle \Phi \rangle_0$ that embeds the test state $\mathbf{u}_\text{test}$ with eq. \ref{eq:loss}.
    \item Set $g=0$ upon this instance. Commence the closed-loop prediction as outlined in sect. \ref{sss:prediction}.
\end{enumerate}

We remind that the method outlined above all takes place during the warm-up phase---that is, the reservoir still runs in an open loop, where the output of the reservoir does not feed back as the input for the next timestep. We also note that extrapolation curve during the open-loop stage is different from that during the closed-loop stage. In the former case, an input of one state is continuously sent into the network, which has the effect of an external force that tries to realign the network behavior to output the same state as the input. This prevents the network from extrapolating to a wider range of states in the neighborhood of the input, whereas in the closed-loop case, a small deviation of the output at one timestep can amplify during the subsequent loops, thereby providing a large enough force for the network to hop into a basin of attraction of a state $\lambda$ where $\lambda-\lambda_{\text{input}}$ is much larger. 

\subsection{Data availability}

The dataset of the Thomas system used in this study is provided on GitHub\cite{github}, and the equations and information necessary to generate the other simulated data are provided in the manuscript.

\subsection{Code availability}

The algorithm code utilized in the current study is available on GitHub\cite{github}.

\bibliographystyle{Science_new1}
\bibliography{scibib} 

\providecommand{\noopsort}[1]{}\providecommand{\singleletter}[1]{#1}%
\begin{thebibliography}{10}

\bibitem{backprop}
Tyagi, K., Rane, C., Manry, M., {\it Artificial Intelligence and Machine Learning for EDGE Computing\/}, Pandey, R., Khatri, S.~K., kumar Singh, N., Verma, P., eds. (Academic Press, 2022), pp. 3--22.

\bibitem{hebbian}
Siri, B., Berry, H., Cessac, B., Delord, B., Quoy, M., A Mathematical Analysis of the Effects of Hebbian Learning Rules on the Dynamics and Structure of Discrete-Time Random Recurrent Neural Networks, {\it Neural Computation\/} {\bf 20}, 2937 (2008).

\bibitem{catastrophic1}
French, R.~M., Catastrophic forgetting in connectionist networks, {\it Trends in Cognitive Sciences\/} {\bf 3}, 128 (1999).

\bibitem{catastrophic2}
McCloskey, M., Cohen, N., Catastrophic Interference in Connectionist Networks: The Sequential Learning Problem, {\it Psychology of Learning and Motivation - Advances in Research and Theory\/} {\bf 24}, 109 (1989).

\bibitem{ucar}
Ucar, H., {\it et~al.\/}, Mechanical actions of dendritic-spine enlargement on presynaptic exocytosis, {\it Nature\/} {\bf 600}, 686 (2021).

\bibitem{qixin_elife}
Yang, Q., {\it et~al.\/}, Cortical waves mediate the cellular response to electric fields, {\it eLife\/} {\bf 11}, e73198 (2022).

\bibitem{qixin_pnas}
Yang, Q., {\it et~al.\/}, Nanotopography modulates intracellular excitable systems through cytoskeleton actuation, {\it Proceedings of the National Academy of Sciences\/} {\bf 120}, e2218906120 (2023).

\bibitem{abby}
Bull, A.~L., {\it et~al.\/}, Actin Dynamics as a Multiscale Integrator of Cellular Guidance Cues, {\it Frontiers in Cell and Developmental Biology\/} {\bf 10} (2022).

\bibitem{actincalciumosc}
Wu, M., Wu, X., Camilli, P.~D., Calcium oscillations-coupled conversion of actin travelling waves to standing oscillations, {\it Proceedings of the National Academy of Sciences\/} {\bf 110}, 1339 (2013).

\bibitem{kate}
O'Neill, K.~M., {\it et~al.\/}, Decoding {Natural} {Astrocyte} {Rhythms}: {Dynamic} {Actin} {Waves} {Result} from {Environmental} {Sensing} by {Primary} {Rodent} {Astrocytes}, {\it Advanced Biology\/} {\bf 7}, e2200269 (2023).

\bibitem{thomas}
Thomas, R., Deterministic chaos seen in terms of feedback circuits: analysis, synthesis, "Labyrinth chaos", {\it International Journal of Bifurcation and Chaos\/} {\bf 09}, 1889 (1999).

\bibitem{ddeimmune}
Rihan, F., {Abdel Rahman}, D., Lakshmanan, S., Alkhajeh, A., A time delay model of tumour–immune system interactions: Global dynamics, parameter estimation, sensitivity analysis, {\it Applied Mathematics and Computation\/} {\bf 232}, 606 (2014).

\bibitem{ddenervous}
Lu, Q., Wang, Q., Shi, X., {\it Delay {Differential} {Equations}: {Recent} {Advances} and {New} {Directions}\/} (Springer US, Boston, MA, 2009), pp. 1--31.

\bibitem{mackey}
Mackey, M.~C., Glass, L., Oscillation and Chaos in Physiological Control Systems, {\it Science\/} {\bf 197}, 287 (1977).

\bibitem{lorenz}
Lorenz, E.~N., Deterministic Nonperiodic Flow, {\it Journal of Atmospheric Sciences\/} {\bf 20}, 130  (1963).

\bibitem{yorke}
Yorke, J.~A., Yorke, E.~D., Metastable chaos: {The} transition to sustained chaotic behavior in the {Lorenz} model, {\it Journal of Statistical Physics\/} {\bf 21}, 263 (1979).

\bibitem{jaeger_rc}
Jaeger, H., Haas, H., {\it Science\/} {\bf 304}, 78 (2004).

\bibitem{maass_lsm}
Maass, W., Natschläger, T., Markram, H., Real-time computing without stable states, {\it Neural Computation\/} {\bf 14}, 2531 (2002).

\bibitem{chaos}
Lu, Z., {\it et~al.\/}, Reservoir observers: Model-free inference of unmeasured variables in chaotic systems, {\it Chaos\/} {\bf 27}, 041102 (2017).

\bibitem{chaos_neuron}
Sussillo, D., Abbott, L., Generating Coherent Patterns of Activity from Chaotic Neural Networks, {\it Neuron\/} {\bf 63} (2009).

\bibitem{sakaguchi}
Sakaguchi, H., Shinomoto, S., Kuramoto, Y., Mutual Entrainment in Oscillator Lattices with Nonvariational Type Interaction, {\it Progress of Theoretical Physics\/} {\bf 79}, 1069 (1988).

\bibitem{jaeger_esn}
Jaeger, H., The ``echo state" approach to analysing and training recurrent neural networks, {\it GMD Report 148\/} (2001).

\bibitem{coordinatechange}
Miller, K.~D., Fumarola, F., Mathematical equivalence of two common forms of firing rate models of neural networks, {\it Neural Computation\/} {\bf 24}, 25 (2012).

\bibitem{kuramoto}
Kuramoto, Y., {\it Chemical {Oscillations}, {Waves}, and {Turbulence}\/}, vol.~19 of {\it Springer {Series} in {Synergetics}\/} (Springer, Berlin, Heidelberg, 1984).

\bibitem{github}
Kang, H., rhythmic\_sharing, GitHub repository, \url{https://github.com/kangh1/rhythmic_sharing} (2025).

\end{thebibliography}
\section{Acknowledgements}
This material is based upon work supported by the Air Force Office of Scientific Research under award number FA9550-21-1-0352 and the Army Research Laboratory cooperative agreement W911NF2320040. The authors would like to thank Nicholas Mennona, Anna M. Emenheiser, Sylvester J. Gates III, Walter Peregrim, Karima Jeneh Perry, Noah Chongsiriwatana, and Kate M. O'Neill for discussions.

\section{Author contributions}
H.K. and W.L. jointly conceived the learning paradigm. H.K. developed and wrote the algorithm used in the manuscript. H.K. and W.L. have jointly written the manuscript.

\section{Competing interests}
The authors declare no competing interests.

\end{document}